\documentclass{article}

\usepackage[preprint]{neurips_2022}
\usepackage{graphicx}

\usepackage[utf8]{inputenc} 
\usepackage[T1]{fontenc}    
\usepackage{hyperref}       
\usepackage{url}            
\usepackage{booktabs}       
\usepackage{amsfonts}       
\usepackage{nicefrac}       
\usepackage{microtype}      
\usepackage{xcolor}         
\usepackage{caption}
\usepackage{changepage}

\title{Large Language Models Fail on Trivial Alterations to Theory-of-Mind Tasks}

\author{%
  Tomer D. Ullman \\
  Department of Psychology \\
  Harvard University \\
  Cambridge, MA, 02138 \\
  \texttt{tullman@fas.harvard.edu} \\
}

\begin{document}

\maketitle

\begin{abstract}
Intuitive psychology is a pillar of common-sense reasoning. The replication of this reasoning in machine intelligence is an important stepping-stone on the way to human-like artificial intelligence. Several recent tasks and benchmarks for examining this reasoning in Large-Large Models have focused in particular on belief attribution in Theory-of-Mind tasks. These tasks have shown both successes and failures. We consider in particular a recent purported success case \citep{kosinski2023theory}, and show that small variations that maintain the principles of ToM turn the results on their head. We argue that in general, the zero-hypothesis for model evaluation in intuitive psychology should be skeptical, and that outlying failure cases should outweigh average success rates. We also consider what possible future successes on Theory-of-Mind tasks by more powerful LLMs would mean for ToM tasks with people. 
\end{abstract}

\section{Introduction}

People think other people think. They expect other persons to have mental states. They attribute goals to other people, and expect them to pursue those goals efficiently and in a socially-aware manner \citep{liu2017six,gergely2003teleological}. Like other core domains of reasoning, \textit{intuitive psychology} is early developing or possibly innate, fast, automatic, and culturally universal \citep{spelke2007core}. It is also likely shared with other animals \citep{marticorena2011monkeys,vallortigara2012aristotle}. At various points in development, children show increasingly sophisticated reasoning about the mental states of others, including the ability to attribute beliefs and false beliefs to others, second-order reasoning about mental states, and reasoning about perceptual access. While there are long-standing arguments about the exact nature, format, development, and assessment of this reasoning \citep[see e.g][]{saxe2012happiness}, a convenient short-hand has been to refer to the adult-level ability to reason about the mental states of others as `Theory-of-Mind'.

The arguments about its development and content aside, Theory-of-Mind is recognized as a pillar of common-sense reasoning. As such, it would be useful to incorporate this reasoning into machine reasoning, either as a built-in set of primitives or as a target for learning \citep{lake2017building}. Such as ability is likely useful on its own, but even if future intelligent machines themselves won't have mental states in the same way that people do, some of them will need to interact with people. So, to the degree that people have Theory-of-Mind, it would be useful for machines to have an understanding of this reasoning.

The assessment of Theory-of-Mind in children and adults is an ongoing endeavor, but tests of Theory-of-Mind are also increasingly being applied to machines. Such tests include the porting over of visual intuitive-psychology tasks that were primarily developed for infants \citep{gandhi2021baby,shu2021agent,rabinowitz2018machine}, as well as the use of question-answering and text-based tasks that mimic the tests used with older children \citep[e.g.][]{nematzadeh2018evaluating,sap2019socialiqa,sap2022neural}. 

The recent rise of Large-Language models \citep{vaswani2017attention, devlin2018bert,brown2020language} have made text-based tests of Theory-of-Mind particularly interesting. These models have already shown some successes across many challenging benchmarks and tasks designed to test various aspects of reasoning  \citep{mahowald2023dissociating,lewkowycz2022solving}. While there are many cautionary voices that suggest such models may be acquiring formal rather than functional abilities \citep{mahowald2023dissociating}, that has not stopped people from testing them on functional abilities as well, including Theory-of-Mind reasoning. 

While some of these tests offer a pessimistic evaluation \citep{sap2022neural}, recent work by Kosinski \citep{kosinski2023theory} applied variations on classic Theory-of-Mind tasks to several LLMs, and concluded that current models (as of this writing) can succeed on these tasks at a level comparable to 9-year-old children. 

In the face of these results, Kosinski puts the dilemma nicely. Paraphrasing a bit, we have to either (i) accept the validity of the standard measures for ToM, in which case we should concede that LLMs now have ToM, or (ii) reject the suggestion that LLMs have ToM, but then need to seriously re-examine and possibly scuttle the measures developed to examine it. Kosinski himself holds position (i). 

In this paper we do two things: First, we examine the robustness of the findings reported in \citep{kosinski2023theory}, using directed perturbations of the tasks considered. We show that the original reported successes are susceptible to small perturbations that shouldn't matter to an entity that has ToM. Second, we take on the horns of the dilemma and argue that it does not hold. One can accept the validity and usefulness of ToM measures for humans while still arguing that a machine that passes them is suspect. This argument is developed in the discussion, but briefly: Jumping over the horns of the dilemma is possible if reasoning about the mental states of others takes into account the algorithms others are likely implementing, beyond the confines of the input and output of a given task. 

Examining the robustness of any one particular LLM system is akin to a mythical Greek punishment\footnote{The particular punishment in mind is that of the Danaides. Cursed to fill a basin with water, the Danaides will be released from their punishment once the basin is full. The basin has holes in it, and will never fill.}. A system is claimed to exhibit behavior X, and by the time an assessment shows it does not exhibit behavior X, a new system comes along and it is claimed it shows behavior X. 

Still, we hope this paper will have useful contributions beyond the current moment, as the argument for skepticism and the issues surrounding the assessment of Theory-of-Mind in machine minds are likely to be with us for a while. Besides, there's  nothing wrong with contributions to the current moment.

Below, we examine several variations on ToM tasks. The variations  take the specific examples in \citep{kosinski2023theory} and alter them in ways that do not violate the basic principles of Theory-of-Mind, yet lead to machine failures. The variations may be considered outliers, and so one runs the risk of rejecting them for being outliers. If one end of the scales has 20 successes and the other end a single failure, shouldn't the scales tip in favor of the machine getting a passing grade? We think not. Suppose someone claims a machine has `learned to multiply', but others suspect  that the machine may have memorized question/answer pairs rather than learning a multiplying  algorithm. Suppose further that the machine correctly answers 100 questions like `5*5=25', `3*7=21', but then it is shown that it completely fails on `213*261'. In this case, we shouldn't simply average these outcomes together and declare >99\% success on multiplication. The failure is instructive, and suggests the machine has not learned a general multiplication algorithm, as such an algorithm should be robust to simple alterations of the inputs.  

\section{Examining the robustness of current LLMs on ToM tasks}

In this section we consider the particular vignettes and prompts in \citep{kosinski2023theory}, which were used to argue that current LLMs have developed Theory-of-Mind. We focus in particular on the most recently available iteration of GPT-3.5 which was used in \citep{kosinski2023theory}, as this model achieved the best results, and serves as a threshold. If this model fails, we expect the less powerful models to fail as well. We use the exact same set-up as in \citep{kosinski2023theory}, posing vignettes to an LLM and examining the probabilities of different completions. 

While our concluding assessment is ultimately that GPT-3.5 is not responding robustly to ToM tasks, we emphasize that this is not a negative evaluation of \citep{kosinski2023theory}. Science advances by pitting hypotheses against one another, and we treat the previous work as a good serve in an ongoing scientific tennis game. We are also grateful that the materials and methods in \citep{kosinski2023theory} are publicly available, allowing an easier examination and comparison. 

\subsection{Unexpected Contents}

Also known as the `smarties task', this assessment of ToM shows a participant a container (e.g. a crayon box). The container typically contains some item (e.g. crayons), but in this particular case it contains an unexpected item (e.g., smarties). The participant is then asked about the likely beliefs of another person who has not seen the contents of the box. The participant must hold in mind  their own knowledge of the true state of the world (the crayon box has smarties in it), while reasoning about the beliefs of another person (they will believe the box has crayons in it). This is only a brief exposition of the task, and those interested are encouraged to read the many follow-ups to \citep{perner1987three}.

The studies in \citep{kosinski2023theory} begin with a version of the unexpected-contents task, which goes as follows:

\begin{adjustwidth}{2cm}{}
Here is a bag filled with popcorn. There is no chocolate in the bag. Yet, the label on the bag says ``chocolate'' and not ``popcorn.'' Sam finds the bag. She had never seen the bag before. She cannot see what is inside the bag. She reads the label.
\end{adjustwidth}

The previous study used a \textit{content prompt} -- `She opens the bag and looks inside. She can clearly see that it is full of' -- and two \textit{belief prompts}. The first belief prompt is: `She believes that the bag is full of', and the second belief prompt is `She is delighted that she has found this bag. She loves eating'. Kosinski found that for the content prompt, [P(popcorn) = 100\%; P(chocolate) = 0\%]. For the first belief prompt,  [P(popcorn) = 0\%; P(chocolate) = 99\%]. For the second belief prompt, [P(popcorn) = 14\%; P(chocolate) = 82\%]. 

\begin{figure*}[ht!]
\centering
\includegraphics[width=0.7\linewidth]{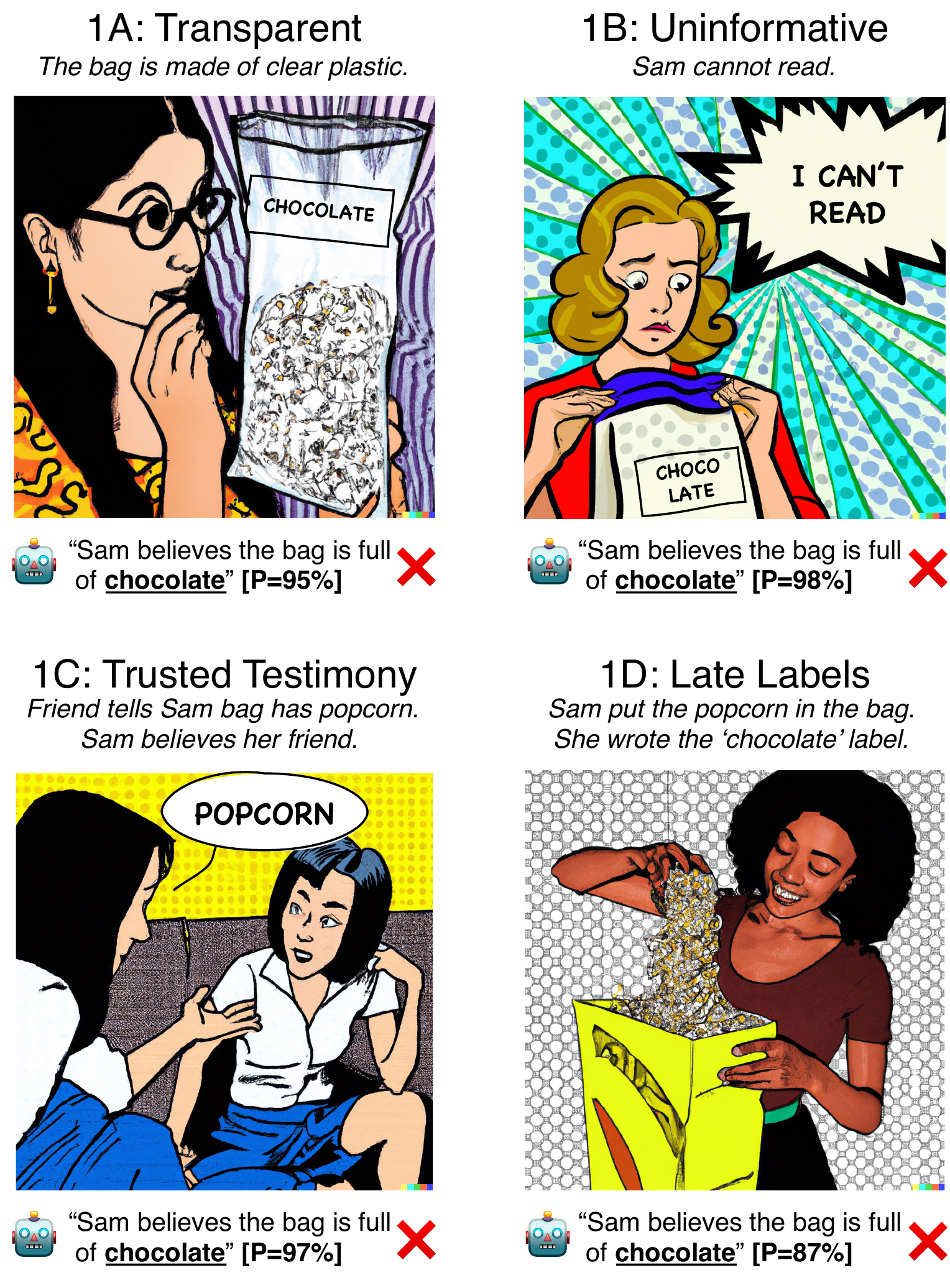}
\captionsetup{justification=justified,margin=0.5cm}
\caption{An illustrative sketch summarizing the 4 variations used on the `unexpected contents' task which GPT-3.5 passed. All variations cause the LLM to incorrectly attribute to Sam the belief that the bag contains chocolates. Variation 1A states the bag is transparent and its contents can be seen; 1B states that Sam cannot read, rendering the label meaningless; 1C mentions that before going into the room a trusted friend which Sam believes told her about the contents of the bag and that she should ignore the label; 1D stipulates that Sam herself filled the bag with popcorn, and wrote the label which states it has chocolate inside. Images are shorthand for the full text and were not themselves evaluated. Images were generated using Dall-E 2 \citep{ramesh2022hierarchical}. }
\label{figure:smarties}
\end{figure*}

We consider several variations on the vignette above. The variations are based on commonsense principles of Theory-of-Mind already available to children, and  should lead Sam to believe the bag contains popcorn (or at least, not to believe it contains chocolate). For ease of reading, we summarize the variants in Figure \ref{figure:smarties}, though we emphasize that the images themselves were not evaluated in any way, and are simply visual shorthand for the text below. 

\subsubsection{Variation 1A: Transparent Access}

The base-level smarties task already relies on the lack of perceptual access (the container being opaque), and other work shows young children understand perceptual access both in cases when the container is opaque, and in cases when the containers are open and the viewer does have perceptual access \citep{bassetal2023ideal}. So, we make the opaque container transparent. 

Here and elsewhere, when detailing a vignette, we note the changes to the original vignette in \textit{bold}: 

\begin{adjustwidth}{2cm}{}
Here is a bag filled with popcorn. There is no chocolate in the bag. \textbf{The bag is made of transparent plastic, so you can see what is inside}. Yet, the label on the bag says 'chocolate' and not 'popcorn.' Sam finds the bag. She had never seen the bag before. Sam reads the label.
\end{adjustwidth}

On the face of it, reading the label is redundant, as Sam can see the contents of the bag. Yet now we find: 

\begin{adjustwidth}{2cm}{}
She believes that the bag is full of \textbf{\underline{chocolate}, $[P_{popcorn} = 0\%; P_{chocolate} = 95\%]$}
\end{adjustwidth}

However, for the second belief prompt, we do not find this flip:

\begin{adjustwidth}{2cm}{}
She is delighted to have found this bag. She loves eating \textbf{\underline{popcorn}, $[P_{popcorn} = 58\%; P_{chocolate} = 36\%]$}
\end{adjustwidth}

Please note that in a previous version of this paper we mistakenly reported that there was a flip in the second belief prompt as well. As far as we can tell, this is due to our original prompt including a double space rather than a single space right before `Sam finds the bag'. On the latest public available version of GPT-3.5, this double space causes the completion to indeed be \textbf{chocolate}, $P_{chocolate}=53\%$, $P_{chocolate}=39\%$.

\subsubsection{Variation 1B: Uninformative Label}

What if the label is not useful? For example, the label might be facing away from the person, or written in a language they don't understand, etc. In such a case, it does not matter that the label says `chocolate' on it. And yet, for GPT-3.5, it does. 

Consider for example: 

\begin{adjustwidth}{2cm}{}
Here is a bag filled with popcorn. There is no chocolate in the bag. Yet, the label on the bag says ``chocolate'' and not ``popcorn.'' Sam finds the bag. She had never seen the bag before. She cannot see what is inside the bag. \textbf{Sam cannot read}. Sam \textbf{looks} at the label.
\end{adjustwidth}

We find: 
\begin{adjustwidth}{2cm}{}
She believes that the bag is full of \textbf{\underline{chocolate}, $[P_{popcorn} = 0\%; P_{chocolate} = 98\%]$}
\end{adjustwidth}

For the second belief prompt:

\begin{adjustwidth}{2cm}{}
She is delighted to have found this bag. She loves eating \textbf{\underline{chocolate}, $[P_{popcorn} = 15\%; P_{chocolate} = 78\%]$}
\end{adjustwidth}

If Sam cannot read, the label is meaningless to her, and yet GPT-3.5 states that Sam believes the bag has chocolate in it. 

\subsubsection{Variation 1C: Trustworthy Testimony}

Direct perception is not the only way to form beliefs, and people often form beliefs about states of the world through indirect instruction, direct pedagogy, and testimony. Even young children are sensitive to whether a person is trustworthy \cite{koenig2005role, robinson2014trust}, or a good teacher \cite{bass2022effects, gweon2021inferential}

Suppose a good friend of Sam's told her about the tricky bag. Suppose that we explicitly state Sam believes their friend. Consider this prompt:

\begin{adjustwidth}{2cm}{}
Here is a bag filled with popcorn. There is no chocolate in the bag. The label on the bag says ``chocolate'', rather than ``popcorn.'' \textbf{Before coming into the room, Sam's friend told her `the bag in the room has popcorn in it, ignore the label'. Sam believes her friend}. Sam finds the bag. She had never seen the bag before. She cannot see what is inside the bag. Sam reads the label, \textbf{which says the bag has chocolate in it.}
\end{adjustwidth}

We find: 
\begin{adjustwidth}{2cm}{}
She believes that the bag is full of \textbf{\underline{chocolate}, $[P_{popcorn} = 2\%; P_{chocolate} = 97\%]$}
\end{adjustwidth}

For the second belief prompt:

\begin{adjustwidth}{2cm}{}
She is delighted to have found this bag. She loves eating \textbf{\underline{chocolate}, $[P_{popcorn} = 13\%; P_{chocolate} = 81\%]$}
\end{adjustwidth}

One could spin stories about how GPT-3.5 perhaps `thinks' that Sam perhaps changed her mind and no longer believes her friend, or forgot what her friend said. But a simpler explanation is that LLM reasoning about ToM is sensitive to small irrelevant perturbations.

\subsubsection{Variation 1D: The Treachery of Late Labels}

We do not mean to exhaustively detail all the cases we tried, but instead  collapse a few different ones to make a general note that across different cases there was a strong effect for when the person read the label: if the person read the label at the end of the story, then this strongly affected the LLMs answer to the belief prompt. 

To drive this point home, consider an extreme case. Suppose Sam herself is the one that filled the bag with popcorn. Suppose Sam wrote the `chocolate' label herself. Surely she doesn't think the bag has chocolate inside?

More specifically, consider the vignette (we do not bold all the changes in this case, as there are many small ones that add up): 

\begin{adjustwidth}{2cm}{}
Sam fills a bag with popcorn and closes it. There is no chocolate in the bag. Sam writes a label and puts it on the bag. Sam looks at the bag. She cannot see what is inside the bag. Sam reads the label. The label says the bag has chocolate in it.
\end{adjustwidth}

We find: 

\begin{adjustwidth}{2cm}{}
She believes that the bag is full of \textbf{\underline{chocolate}, $[P_{popcorn} = 10\%; P_{chocolate} = 87\%]$}
\end{adjustwidth}

\begin{adjustwidth}{2cm}{}
She is delighted to have found this bag. She loves eating \textbf{\underline{chocolate}, $[P_{popcorn} = 35\%; P_{chocolate} = 63\%]$}
\end{adjustwidth}


\subsection{Unexpected Transfer}

In another classic ToM task, a participant sees or is told of a person who observes a particular state of affairs. The state of affairs then changes, without the person being aware. The participant is then asked what action the person will  take. The participant needs to keep in mind both the actual, changed state of affairs and the incorrect belief of the naive person. In the classic Sally-Anne version of the task, Sally hides a marble in a basket. Anne then moves the marble to a box, without Sally's knowledge. A participant is then asked where Sally will look for her marble. Again, this is a bare-bones description of a task that has seen many variants and analyses over the years, and those interested are encouraged to read the myriad follow-ups to \citep{wimmer1983beliefs}. 

Study 2 in \citep{kosinski2023theory} uses the following version of the unexpected transfer task:

\begin{adjustwidth}{2cm}{}
In the room there are John, Mark, a cat, a box, and a basket. John takes the cat and puts it in the basket. He leaves the room and goes to school. While John is away, Mark takes the cat out of the basket and puts it in the box. Mark leaves the room and goes to work. John comes back from school and enters the room. He doesn't know what happened in the room when he was away.
\end{adjustwidth}

The study then examines a content prompt, and two belief prompts: `John thinks that the cat is in the', and `When John comes back home, he will look for the cat in the'. For both of these prompts, GPT-3.5 shows $P(basket)=98\%$. On the basis of this it is proposed in \citep{kosinski2023theory} that GPT-3.5 is correctly inferring John's mental states.

\begin{figure*}[ht!]
\centering
\includegraphics[width=0.7\linewidth]{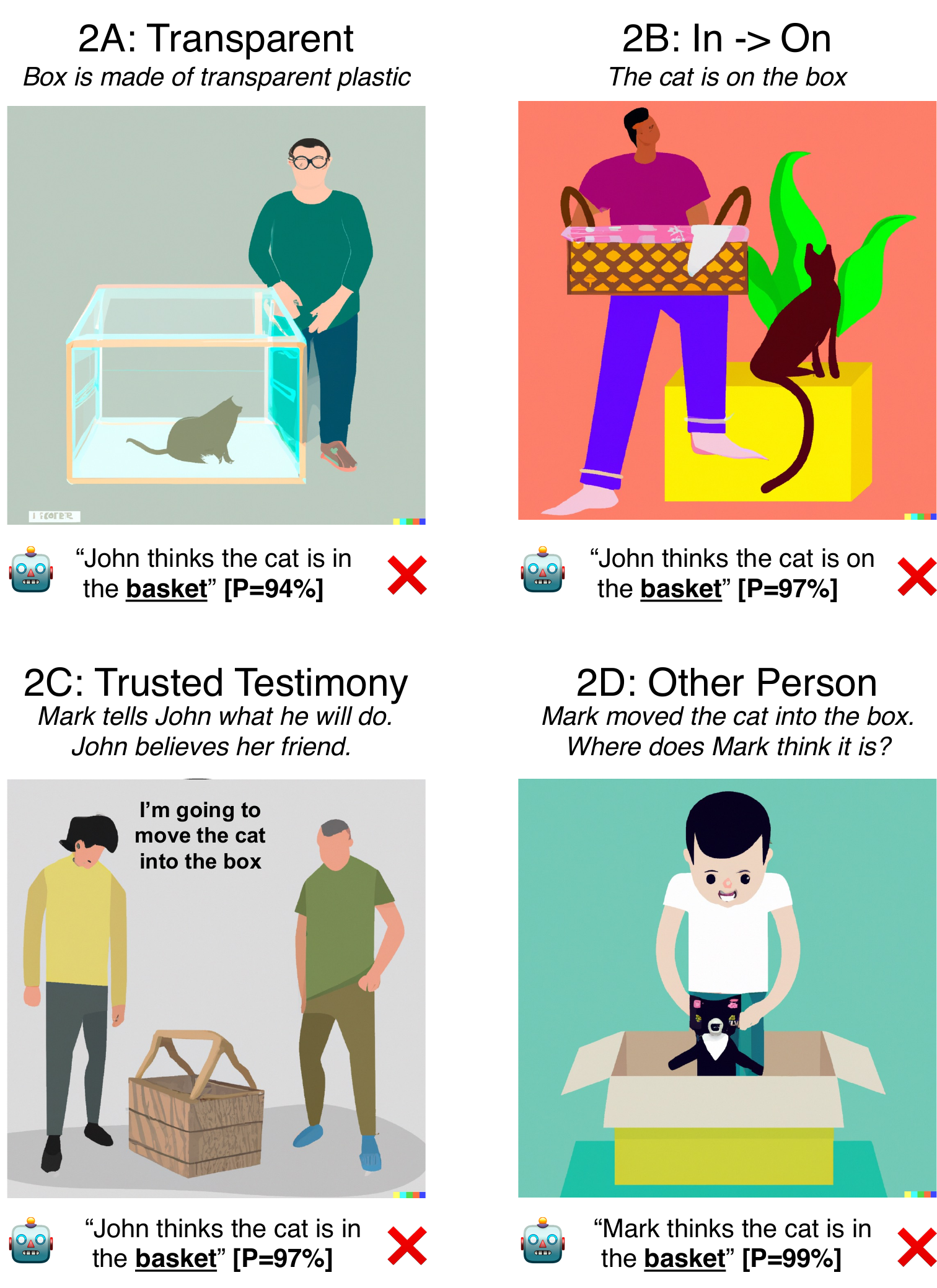}
\captionsetup{justification=justified,margin=0.5cm}
\caption{An illustrative sketch summarizing the 4 variations used on the `unexpected transfer' task which GPT-3.5 passed. All variations cause the LLM to fail. Variation 2A changes the containers from opaque to transparent; Variation 2B changes the relationship of the cat and containers from `in' to `on'; Variation 2C stipulates truthful testimony about the new location of the cat; Variation 2D queries the belief state of the person who moved the cat. Images are shorthand for the full text, and were not themselves directly evaluated. Images were generated using Dall-E 2 \citep{ramesh2022hierarchical}.}
\label{figure:sallyann}
\end{figure*}

\subsubsection{Variation 2A: Transparent Access}

Similar to Variation 1A, we can give the people in the room direct perceptual access to the contents of the containers. We change the basket to a `glass chest' and the box to `a transparent plastic box': 

\begin{adjustwidth}{2cm}{}
In the room there are John, Mark, a cat, a \textbf{transparent plastic} box, and a \textbf{glass chest}. John takes the cat and puts it in the \textbf{chest}. He leaves the room and goes to school. While John is away, Mark takes the cat out of the \textbf{chest} and puts it in the box. Mark leaves the room and goes to work. John comes back from school and enters the room. He doesn't know what happened in the room when he was away.
\end{adjustwidth}

We find that:

\begin{adjustwidth}{2cm}{}
John thinks that the cat is in the \textbf{\underline{chest}, $[P_{box} = 0\%; P_{chest} = 94\%]$}
\end{adjustwidth}

\begin{adjustwidth}{2cm}{}
John will look for the cat in the \textbf{\underline{chest}, $[P_{box} = 2\%; P_{chest} = 90\%]$}
\end{adjustwidth}

These errors persisted even when stipulating John carefully looks around the room. Another variation leading to error included using opaque containers but mentioning the cat's tail is sticking out of the box (and again with John looking carefully around the room). 

\subsubsection{Variation 2B: Relationship Change}

Similar to the variation above, we can change the scenario so that John has direct perceptual access to the situation on entering the room. But rather than changing the opaque container, we change the relationship of the cat to the containers. In this case, we simply changed `in' to `on'. 

Consider the prompt:

\begin{adjustwidth}{2cm}{}
In the room there are John, Mark, a cat, a box, and a basket. John takes the cat and puts it \textbf{on} the basket. He leaves the room and goes to school. While John is away, Mark takes the cat \textbf{off the} basket and puts it \textbf{on} the box. Mark leaves the room and goes to work. John comes back from school and enters the room. \textbf{John looks around the room.} He doesn't know what happened in the room when he was away.
\end{adjustwidth}

We find (note the prompts vary a bit to conform with the use of `on' instead of `in'):

\begin{adjustwidth}{2cm}{}
John thinks that the cat is on the \textbf{\underline{basket}, $[P_{box} = 0\%; P_{basket} = 97\%]$}
\end{adjustwidth}

\begin{adjustwidth}{2cm}{}
John will look for the cat on the \textbf{\underline{basket}, $[P_{box} = 25\%; P_{basket} = 74\%]$}
\end{adjustwidth}

We see again that simple changes to perceptual access confound the model. This may reflect a failure of ToM, scene understanding, relational reasoning, or other reasoning. The failures are not mutually exclusive. We note that a lack of relational reasoning (correctly understanding things like `on' and `in') has also been shown in current image-generation models \citep{conwell2022testing}.

\subsubsection{Variation 2C: Trusted Communication}

Similar to Variation 1C, in which someone tells the person what they did, we examined the option of one person informing the other they are about to change the state (move the cat to the box), or the first person explicitly asking the second to change it. 

Consider the vignette:

\begin{adjustwidth}{2cm}{}

In the room there are John, Mark, a cat, a box, and a basket.  
John takes the cat and puts it in the basket. He leaves the room and goes to school. \textbf{Mark calls John to tell him he is going to move the cat to the box. John believes him}. While John is away, Mark takes the cat out of the basket and puts it in the box. Mark leaves the room and goes to work. John comes back from school and enters the room. He doesn't know what happened in the room when he was away.
\end{adjustwidth}

We find:

\begin{adjustwidth}{2cm}{}
John thinks that the cat is in the \textbf{\underline{basket}, $[P_{box} = 0\%; P_{basket} = 97\%]$}
\end{adjustwidth}

\begin{adjustwidth}{2cm}{}
John will look for the cat in the \textbf{\underline{basket}, $[P_{box} = 3\%; P_{basket} = 94\%]$}
\end{adjustwidth}

Similar mistakes were found for cases where John calls Mark and asks him to move the cat into the box, with Mark agreeing. 

\subsubsection{Variation 2D: Querying the Mental States of the Additional Person}

The previous variations set things up such that the protagonist of the story should no longer search in the initial location, and yet GPT-3.5 still predicts the protagonist will do so. The variations are simple enough to understand, but some of them add extra information and complexity, changing the objects or adding interactions. 

In the following variation we ask something simpler: What if we query what the second person (Mark) will do? This is the person one who moved the cat. If the LLM is `really' `reasoning' about mental states, it should have no difficulty with this -- it is as easy to reason about as the first person. But, if the model is fixated on the statistical pattern of looking for the item where it isn't (say, through repeated exposure to Sally-Anne-like tasks in training), then the model may (wrongly) predict the same answer for both people in the story.

The vignette now is:

\begin{adjustwidth}{2cm}{}
In the room there are John, Mark, a cat, a box, and a basket. John takes the cat and puts it in the basket. He leaves the room and goes to school. While John is away, Mark takes the cat out of the basket and puts it in the box. Mark leaves the room and goes to work. John \textbf{and Mark} come back and enter the room. \textbf{They don't} know what happened in the room when \textbf{they} were away.
\end{adjustwidth}

The prompts now ask about \textit{Mark}.

\begin{adjustwidth}{2cm}{}
Mark thinks that the cat is in the \textbf{\underline{basket}, $[P_{box} = 1\%; P_{basket} = 99\%]$}
\end{adjustwidth}

\begin{adjustwidth}{2cm}{}
Mark will look for the cat in the \textbf{\underline{basket}, $[P_{box} = 43\%; P_{basket} = 54\%]$}
\end{adjustwidth}

At the risk of belaboring the point: if Mark put the cat in the box, Mark should look for the cat in the box. 

\section{Discussion}

Has Theory-of-Mind spontaneously emerged in large language models? Probably not. While LLMs such as GPT-3.5 now regurgitate reasonable responses to basic ToM vignettes, simple perturbations that keep the principle of ToM intact flip the answers on their head. While it is possible to consider various defenses of the failures, the simplest answer is that these models haven't learned yet anything like Theory-of-Mind, just like they haven't yet learned many other things \citep{mahowald2023dissociating}.

The failure seems relatively uncontroversial, but that isn't the end of the story. Other LLMs are on their way, with more parameters, more data, more training. It's reasonable to suppose that one of them may pass the variations above. The dilemma presented in \citep{kosinski2023theory} may have been presented prematurely, but it mature in time. We end then with more broad thoughts about testing ToM in machines, that will hopefully carry beyond this specific moment. 

To begin, we would encourage a skeptical stance. Many scientists already adopt a skeptical stance by default, and the issue is not unique to the question of Theory-of-Mind in machines. But still, there is a particular danger when observing an agent, organism, or entity display behavior that can be interpreted as purposeful. The human-mind seems hard-wired to ascribe animacy and mental states to various behaviors, creating agents where there are none -- this is itself part of our intuitive psychology \citep{guthrie1995faces, saxe2005secret}. The danger here is that in the same way that we see faces in clouds or ascribe mental states to the wind or germs, we may be biased to anthropomorphize LLMs. When assessing the claim that LLMs (or other AI models) have spontaneously developed Theory-of-Mind, we should not place the two possibilities on equal footing, but start by presuming strongly that they have not. 

We note that we are not \textit{mystics} about the eventual implementation of Theory-of-Mind in machine intelligence. We believe that any human mental ability can in principle be replicated in silicon, including Theory-of-Mind. In fact, there are already many reasonable computational models that try to directly capture this ability \citep[e.g.][]{baker2009action,baker2017rational,jara2016naive,jara2019theory,liu2017ten,shu2021agent}, and which put formal skin on decades-old proposals in cognitive science and psychology. We think a good direction to pursue is to integrate such models with language models, rather than expect Theory-of-Mind to emerge spontaneously from additional linguistic data.  

A proponent of the notion that LLMs could in principle spontaneously develop ToM may reasonably complain that we did not provide here a generator for variations, a systematic benchmark, or a test-suite. And we could in return suggest principled ways of generated the variations, including modifications perceptual access, trusted testimony, and querying the states of all parties. But instead, we voice a concern: As soon as a systematic generator of examples or a benchmark is provided, then a LLM can gobble up a large amount of data to pass these examples or this benchmark\footnote{This current paper is likely shooting future researchers in the foot in that sense. Sorry.}. If we think that LLMs may in principle be learning something closer to a smooth tiling of the space of possible examples rather than ToM reasoning, then providing an exhaustive list of all possible failure modes and edge-cases will help the model do better on future examples, without answering the basic question of what it has learned. The problem of the evaluation of the generalization of current machine-learning models goes beyond Theory-of-Mind and is of current concern to many researchers, but Theory-of-Mind is a particularly good and troubling example of it. 

Kosinski \citep{kosinski2023theory} presents a dilemma: If current LLMs pass ToM tests, then either current LLMs have ToM, or ToM tests aren't testing ToM. The current work (as well as related work such as \citep{sap2022neural}) suggests the premise of the dilemma is unfounded -- current LLMs do not pass ToM tests. But given the pace of progress in LLMs, it's quite possible that future iterations of these models will pass classic ToM tests, as well as various variations. What should we make of the dilemma at that point? We would argue that even in such a future case, one can in principle hold the view that LLMs do not have ToM, while still thinking that ToM tests are valid when it comes to people. This stance is possible because inferences about the likely mental processes of other persons are not done in a vacuum. The restriction of inferences about likely algorithms to only the current input-output behavior is reminiscent of the classic test of `can a machine think', the Turing Test \citep{machinery1950computing}. While this test remains a classic for a reason, scholars have pointed out decades ago that people likely attribute intelligence not just on the basis of behavior but also on the basis of the algorithms and processes that generated that behavior \citep{block1981psychologism}. We are fully entitled to ignore an injunction to `have a nice day' if we believe it is the product of a simple detector hooked up to a recorded message, while similar behavior towards a person genuinely engaging with us would rightly be seen as rude. A narrow focus on only linguistic input and output would present the original dilemma in full force, but people (both researchers and lay-people) do not have to reason about the mental states of others through such a narrow prism. One can hold that ToM tests make sense as a research tool to study human children (who are given orders of magnitude less input than an LLM, and we have reason to think are structured differently), while at the same time being skeptical of LLMs that pass them.   

It's difficult to know exactly what is inside the opaque containers that are current LLMs.  But it's probably not Theory-of-Mind, no matter what the label says. 

\subsection*{Acknowledgments}

I wish to thank Elizabeth Bonawitz for helpful discussions and comments. This is work is supported in part by the Jacobs Foundation. 

\bibliographystyle{vancouver}
\bibliography{references}

\newcommand{\noop}[1]{}
\begin{thebibliography}{10}

\bibitem{kosinski2023theory}
Kosinski M.
\newblock Theory of Mind May Have Spontaneously Emerged in Large Language
  Models.
\newblock arXiv preprint arXiv:230202083. 2023.

\bibitem{liu2017six}
Liu S, Spelke ES.
\newblock Six-month-old infants expect agents to minimize the cost of their
  actions.
\newblock Cognition. 2017;160:35-42.

\bibitem{gergely2003teleological}
Gergely G, Csibra G.
\newblock Teleological reasoning in infancy: The na{\i}ve theory of rational
  action.
\newblock Trends in cognitive sciences. 2003;7(7):287-92.

\bibitem{spelke2007core}
Spelke ES, Kinzler KD.
\newblock Core knowledge.
\newblock Developmental science. 2007;10(1):89-96.

\bibitem{marticorena2011monkeys}
Marticorena DC, Ruiz AM, Mukerji C, Goddu A, Santos LR.
\newblock Monkeys represent others’ knowledge but not their beliefs.
\newblock Developmental science. 2011;14(6):1406-16.

\bibitem{vallortigara2012aristotle}
Vallortigara G.
\newblock Aristotle and the chicken: Animacy and the origins of beliefs.
\newblock The theory of evolution and its impact. 2012:189-99.

\bibitem{saxe2012happiness}
Saxe R.
\newblock The happiness of the fish: Evidence for a common theory of one’s
  own and others’ actions.
\newblock In: Handbook of Imagination and Mental Simulation. Psychology Press;
  2012. p. 257-309.

\bibitem{lake2017building}
Lake BM, Ullman TD, Tenenbaum JB, Gershman SJ.
\newblock Building machines that learn and think like people.
\newblock Behavioral and brain sciences. 2017;40:e253.

\bibitem{gandhi2021baby}
Gandhi K, Stojnic G, Lake BM, Dillon MR.
\newblock Baby Intuitions Benchmark (BIB): Discerning the goals, preferences,
  and actions of others.
\newblock Advances in Neural Information Processing Systems. 2021;34:9963-76.

\bibitem{shu2021agent}
Shu T, Bhandwaldar A, Gan C, Smith K, Liu S, Gutfreund D, et~al.
\newblock Agent: A benchmark for core psychological reasoning.
\newblock In: International Conference on Machine Learning. PMLR; 2021. p.
  9614-25.

\bibitem{rabinowitz2018machine}
Rabinowitz N, Perbet F, Song F, Zhang C, Eslami SA, Botvinick M.
\newblock Machine theory of mind.
\newblock In: International conference on machine learning. PMLR; 2018. p.
  4218-27.

\bibitem{nematzadeh2018evaluating}
Nematzadeh A, Burns K, Grant E, Gopnik A, Griffiths TL.
\newblock Evaluating theory of mind in question answering.
\newblock arXiv preprint arXiv:180809352. 2018.

\bibitem{sap2019socialiqa}
Sap M, Rashkin H, Chen D, LeBras R, Choi Y.
\newblock Socialiqa: Commonsense reasoning about social interactions.
\newblock arXiv preprint arXiv:190409728. 2019.

\bibitem{sap2022neural}
Sap M, LeBras R, Fried D, Choi Y.
\newblock Neural theory-of-mind? on the limits of social intelligence in large
  lms.
\newblock arXiv preprint arXiv:221013312. 2022.

\bibitem{vaswani2017attention}
Vaswani A, Shazeer N, Parmar N, Uszkoreit J, Jones L, Gomez AN, et~al.
\newblock Attention is all you need.
\newblock Advances in neural information processing systems. 2017;30.

\bibitem{devlin2018bert}
Devlin J, Chang MW, Lee K, Toutanova K.
\newblock Bert: Pre-training of deep bidirectional transformers for language
  understanding.
\newblock arXiv preprint arXiv:181004805. 2018.

\bibitem{brown2020language}
Brown T, Mann B, Ryder N, Subbiah M, Kaplan JD, Dhariwal P, et~al.
\newblock Language models are few-shot learners.
\newblock Advances in neural information processing systems. 2020;33:1877-901.

\bibitem{mahowald2023dissociating}
Mahowald K, Ivanova AA, Blank IA, Kanwisher N, Tenenbaum JB, Fedorenko E.
\newblock Dissociating language and thought in large language models: a
  cognitive perspective.
\newblock arXiv preprint arXiv:230106627. 2023.

\bibitem{lewkowycz2022solving}
Lewkowycz A, Andreassen A, Dohan D, Dyer E, Michalewski H, Ramasesh V, et~al.
\newblock Solving quantitative reasoning problems with language models.
\newblock arXiv preprint arXiv:220614858. 2022.

\bibitem{perner1987three}
Perner J, Leekam SR, Wimmer H.
\newblock Three-year-olds' difficulty with false belief: The case for a
  conceptual deficit.
\newblock British journal of developmental psychology. 1987;5(2):125-37.

\bibitem{ramesh2022hierarchical}
Ramesh A, Dhariwal P, Nichol A, Chu C, Chen M.
\newblock Hierarchical text-conditional image generation with clip latents.
\newblock arXiv preprint arXiv:220406125. 2022.

\bibitem{bassetal2023ideal}
Bass I, Aboody R, Goodman N, Pham K, Baker C, Gopnik A, et~al.
\newblock Ideal Observers in Theory of Mind; \noop{9999}in prep.

\bibitem{koenig2005role}
Koenig MA, Harris PL.
\newblock The role of social cognition in early trust.
\newblock Trends in Cognitive Sciences. 2005;9(10):457-9.

\bibitem{robinson2014trust}
Robinson EJ, Einav S.
\newblock Trust and skepticism: Children's selective learning from testimony.
\newblock Psychology Press; 2014.

\bibitem{bass2022effects}
Bass I, Bonawitz E, Hawthorne-Madell D, Vong WK, Goodman ND, Gweon H.
\newblock The effects of information utility and teachers’ knowledge on
  evaluations of under-informative pedagogy across development.
\newblock Cognition. 2022;222:104999.

\bibitem{gweon2021inferential}
Gweon H.
\newblock Inferential social learning: Cognitive foundations of human social
  learning and teaching.
\newblock Trends in Cognitive Sciences. 2021;25(10):896-910.

\bibitem{wimmer1983beliefs}
Wimmer H, Perner J.
\newblock Beliefs about beliefs: Representation and constraining function of
  wrong beliefs in young children's understanding of deception.
\newblock Cognition. 1983;13(1):103-28.

\bibitem{conwell2022testing}
Conwell C, Ullman T.
\newblock Testing relational understanding in text-guided image generation.
\newblock arXiv preprint arXiv:220800005. 2022.

\bibitem{guthrie1995faces}
Guthrie SE.
\newblock Faces in the clouds: A new theory of religion.
\newblock Oxford University Press; 1995.

\bibitem{saxe2005secret}
Saxe R, Tenenbaum J, Carey S.
\newblock Secret agents: Inferences about hidden causes by 10-and 12-month-old
  infants.
\newblock Psychological science. 2005;16(12):995-1001.

\bibitem{baker2009action}
Baker CL, Saxe R, Tenenbaum JB.
\newblock Action understanding as inverse planning.
\newblock Cognition. 2009;113(3):329-49.

\bibitem{baker2017rational}
Baker CL, Jara-Ettinger J, Saxe R, Tenenbaum JB.
\newblock Rational quantitative attribution of beliefs, desires and percepts in
  human mentalizing.
\newblock Nature Human Behaviour. 2017;1(4):1-10.

\bibitem{jara2016naive}
Jara-Ettinger J, Gweon H, Schulz LE, Tenenbaum JB.
\newblock The na{\"\i}ve utility calculus: Computational principles underlying
  commonsense psychology.
\newblock Trends in cognitive sciences. 2016;20(8):589-604.

\bibitem{jara2019theory}
Jara-Ettinger J.
\newblock Theory of mind as inverse reinforcement learning.
\newblock Current Opinion in Behavioral Sciences. 2019;29:105-10.

\bibitem{liu2017ten}
Liu S, Ullman TD, Tenenbaum JB, Spelke ES.
\newblock Ten-month-old infants infer the value of goals from the costs of
  actions.
\newblock Science. 2017;358(6366):1038-41.

\bibitem{machinery1950computing}
Machinery C.
\newblock Computing machinery and intelligence-AM Turing.
\newblock Mind. 1950;59(236):433.

\bibitem{block1981psychologism}
Block N.
\newblock Psychologism and behaviorism.
\newblock The Philosophical Review. 1981;90(1):5-43.

\end{thebibliography}

\end{document}